\documentclass[letterpaper]{article} 
\usepackage{aaai24}  
\usepackage{times}  
\usepackage{helvet}  
\usepackage{courier}  
\usepackage[hyphens]{url}  
\usepackage{graphicx} 
\usepackage{natbib}  
\usepackage{caption} 
\usepackage{algorithm}
\usepackage{algorithmic}
\usepackage{amsfonts}

\usepackage{xcolor}

\usepackage{multirow}
\usepackage{amsmath}
\usepackage{booktabs}
\usepackage{siunitx}

\newcommand{\dummyhref}[2]{{#2}}

\newcommand{\argmax}{\operatornamewithlimits{arg\,max}}  

\usepackage{newfloat}
\usepackage{listings}
\DeclareCaptionStyle{ruled}{labelfont=normalfont,labelsep=colon,strut=off} 
\floatstyle{ruled}
\newfloat{listing}{tb}{lst}{}
\floatname{listing}{Listing}
\title{Decentralized Monte Carlo Tree Search for Partially Observable Multi-agent Pathfinding}

\author {
    Alexey Skrynnik\textsuperscript{\rm 1,2}, 
    Anton Andreychuk\textsuperscript{\rm 1},
    Konstantin Yakovlev\textsuperscript{\rm 2,1},
    Aleksandr Panov\textsuperscript{\rm 1,2}
}

\affiliations {
    \textsuperscript{\rm 1}AIRI, Moscow, Russia\\
    \textsuperscript{\rm 2} Federal Research Center for Computer Science and Control of Russian Academy of Sciences, Moscow, Russia\\
    \dummyhref{mailto:skrynnikalexey@gmail.com}{skrynnikalexey@gmail.com}, \dummyhref{mailto:andreychuk@airi.net}{andreychuk@airi.net},
    \dummyhref{mailto:yakovlev@isa.ru}{yakovlev@isa.ru},
    \dummyhref{mailto:panov@airi.net}{panov@airi.net}
}

\begin{document}

\maketitle

\begin{abstract}
  The Multi-Agent Pathfinding (MAPF) problem involves finding a set of conflict-free paths for a group of agents confined to a graph. In typical MAPF scenarios, the graph and the agents' starting and ending vertices are known beforehand, allowing the use of centralized planning algorithms. However, in this study, we focus on the decentralized MAPF setting, where the agents may observe the other agents only locally and are restricted in communications with each other. Specifically, we investigate the lifelong variant of MAPF, where new goals are continually assigned to the agents upon completion of previous ones. Drawing inspiration from the successful AlphaZero approach, we propose a decentralized multi-agent Monte Carlo Tree Search (MCTS) method for MAPF tasks. Our approach utilizes the agent's observations to recreate the intrinsic Markov decision process, which is then used for planning with a tailored for multi-agent tasks version of neural MCTS. The experimental results show that our approach outperforms state-of-the-art learnable MAPF solvers. The source code is available at https://github.com/AIRI-Institute/mats-lp.
\end{abstract}

\section{Introduction}
\label{sec:intro}

Multi-agent pathfinding (MAPF) is a non-trivial problem inspired by numerous practical applications like automated warehouses, video games, intelligent transport systems, etc. A large body of works~\cite{ma2019searching,sharon2015conflict,Wagner2011} study this problem in a centralized setting, i.e., it is assumed that a central control unit exists that \emph{i}) has a knowledge of the full state of the environment (locations of the agents, their goals, positions of the static obstacles, etc.) at any time moment; \emph{ii}) is in charge of providing conflict-free solutions to MAPF queries. Indeed, various flavors of MAPF problems are studied within this setting, e.g., Classical MAPF~\cite{stern2019multi} when each agent is assigned a unique goal, Colored MAPF~\cite{ma2016optimal} when the agents are split into teams and agents of one team are interchangeable, Anonymous MAPF~\cite{honig2018conflict}, when any agent can pursue any goal, etc. One MAPF variant, that we study in this work, is the Lifelong MAPF (LMAPF). In this setting, the agents must constantly pursue goals (provided externally), i.e., when an agent reaches its goal, it is immediately assigned another one. This setting is motivated by the real-world delivery applications when a group of robots has to constantly deliver some items dispersed in the shared environment, e.g., items of goods in the warehouse, documents in the office building, medicine in the hospital, etc.

One of the ways to solve LMAPF is to adapt existing MAPF solvers to the lifelong setting. One of such recent methods, RHCR~\cite{li2021lifelong}, involves centralized re-planning every $k$ time-steps. Indeed, when $k$ is small and the number of agents is large, the performance of such an approach degrades significantly as it may take too much time for a solver to construct a joint collision-free plan. Bounded-horizon planning can mitigate this issue to a certain extent; indeed, RHCR utilizes this technique. However, this is still limited. 

An appealing orthogonal approach is to solve LMAPF in a distributed fashion, i.e., model it as a decentralized sequential decision-making process when every agent individually decides what action to take at each time step. Most of the state-of-the-art decentralized (L)MAPF solvers are the learnable ones~\cite{sartoretti2019primal,damani2021primal,ma2021distributed,Li2022MultiAgentPF}. However, the performance of these solvers may rely heavily on the dataset of problem instances used at the learning (training) stage. Their performance often drops significantly in setups that are unlike the latter ones. This is a general problem known in machine learning as low generalization. To mitigate this issue, hybrid approaches were proposed that typically include a (search-based) global planner and a (local) learnable policy that is tailored to follow the global plan while resolving the potential inter-agent conflicts~\cite{Wang2020,9340876}. Such approaches also have limitations because, under challenging cases, it is required to move significantly away from the local sub-goal for the agents to disperse in bottlenecks. Learning-based methods may demonstrate low efficiency in this kind of task. Thus, another way of combining the learning-based and search-based approaches is desirable.

This work follows a hybrid search-and-learning approach to create a decentralized MAPF solver. However, our methodology is different from the ones described above. On the one hand, we rely on the (lightweight) learnable policy that can drive an agent toward a goal. On the other hand, to improve the agent's ability to cooperate, we utilize Monte-Carlo Tree Search (MCTS). This powerful technique is usually used for antagonistic game environments and single-agent tasks~\cite{browne2012survey}. In this work, we follow the seminal AlphaGo approach~\cite{Silver2016,silver2017mastering} and design a variant of MCTS that uses the suggested learnable policy to evaluate environmental states and provide action distributions to grow the search tree. This contributes to the effective simulation of the different variants of how the agent and the neighboring agents might behave in the future and focus on the most prominent variants (using the MCTS machinery). As a result, all agents can exhibit (implicit) coordination and successfully solve challenging LMAPF instances (i.e., the ones involving long corridors or tight passages, etc.). From the reinforcement learning (RL) perspective our approach may be attributed as model-based RL, i.e., we rely both on the learnable policy and on the model of the world to, first, simulate different variants of how the world might evolve in response to our action, and, second, to choose the most promising action, based on this simulation process.

In the empirical evaluation, we compare our method, which we dub \textsc{MATS-LP} (Multi-agent Adaptive Tree Search with the Learned Policy), to the state-of-the-art competitors, i.e., Primal2~\cite{damani2021primal} and SCRIMP~\cite{wang_scrimp_2023} and show that it numerous cases \textsc{MATS-LP} notably outperforms them.

\section{Related Works}

Two streams of research are particularly relevant to our work: learnable (lifelong) MAPF methods and utilizing MCTS for multi-agent systems and MAPF in particular. Next, we review both of these domains.

\paragraph{Learnable (L)MAPF Solvers} Among the recent works dedicated to MAPF, one of the first ones that were specifically dedicated to creating a learning-based MAPF solver was~\cite{sartoretti2019primal}. A combination of reinforcement learning and learning from expert demonstrations was used to create a learnable policy called Primal, tailored to solve conventional MAPF problems. Later in~\cite{damani2021primal}, an enhanced version of this solver, Primal2, was introduced. The latter was equipped with special corridor reasoning techniques, aiming at avoiding the deadlocks in narrow corridors, and it supported lifelong MAPF setting (therefore, we choose Primal2 as one of the baselines we compare our method to). Among the other learnable MAPF solvers that use reinforcement learning to obtain a decision-making policy, one can name~\cite{riviere2020glas,Wang2020}. The learnable methods introduced in~\cite{li2020graph,ma2021distributed,Li2022MultiAgentPF} add communication capabilities to the agents, i.e., allow the agents to communicate to resolve deadlocks and avoid congestion. In this work, we compare with one of the most recent communication-based methods, i.e., SCRIMP~\cite{wang_scrimp_2023}. However, it is worth noting that our method does not rely on agent communication.

\paragraph{MCTS for MAPF}
Initially, Monte Carlo Tree Search (MCTS) algorithms demonstrated their effectiveness in competitive games with complete information, such as chess or Go \cite{silver2017mastering}. More recent versions of MCTS utilize deep neural networks to approximate the values of game states instead of relying solely on simulations. These approaches have also shown promising results in single-agent scenarios, where agents can learn a model of the environment and play Atari games \cite{schrittwieser2020mastering, ye2021mastering}. Besides gaming, MCTS methods have found applications in other domains, such as matrix multiplication optimization \cite{fawzi2022discovering} and theorem proving using the Hyper Tree approach \cite{lample2022hypertree}. Additionally, MCTS techniques have demonstrated applicability in robotics \cite{best2019dec, dam2022monte}.

Despite the growing interest in utilizing MCTS for multi-agent tasks, there have been limited applications of MCTS for MAPF. In their work \cite{zerbel2019multiagent}, the authors propose a multi-agent MCTS for Anonymous MAPF in a grid-world environment. Their environment has a dense reward signal (the agent who reached any goal on the map received a reward and ended the episode), and there are no obstacles, making collision avoidance easier. The authors build a separate tree for each agent using a classical algorithm. They then jointly apply the best actions (forming a plan) from the trees in the simulator to receive true scores of the solution and update the trees on that difference. This approach performs well even with a large number of agents.

A recent paper \cite{skrynnik2021hybrid} proposed a more sophisticated approach for multi-agent planning that combines RL and MCTS. The authors suggested a two-part scheme that includes a goal achievement module and a conflict resolution module. The latter was trained using MCTS. The construction of the search tree for each of the agents was also performed independently, and actions for other agents were selected using the currently trained policy. This work used MCTS only during training to train the conflict resolution policy.

\section{Background}

\paragraph{Multi-agent Pathfinding} We rely on commonly-used MAPF assumptions as described in the survey work on this topic~ \cite{stern2019multi}. The timeline is divided into time steps, and a graph $G=(V, E)$ represents the positions of $K$ agents. Each agent can either wait in its current vertex or move to an adjacent one at each time step. We assume that the outcomes of the actions are deterministic and no inaccuracies occur when executing the actions. A sequence of such actions is referred to as a plan. For different agents, two plans are conflict-free if there are no vertex or edge collisions, meaning that agents do not swap vertices simultaneously or occupy the same vertex at the same time step. MAPF problem generally asks to find a set of $K$ plans $Plans={plan_1, plan_2, ..., plan_K}$, s.t. a plan for agent $i$ starts at the predefined start vertex and ends at the predefined goal vertex, and all pairs of plans are conflict-free. In MAPF, it is common to minimize one of the following cost objectives: $SOC=\sum_{i=1}^n cost(plan_i)$ or $makespan = \max_i cost(plan_i)$. Here, $cost(plan_i)$ represents the individual plan's cost, which is the number of time steps taken by agent $i$ to reach its goal.

In this work, we consider the \textit{lifelong} variant of MAPF (LMAPF), where immediately after an agent reaches its goal, it is assigned to another one (via an external assignment procedure) and has to continue moving to a new goal. Thus, LMAPF generally asks to find a set of $K$ initial plans and update each agent’s plan when it reaches the current goal and receives a new one. In extreme cases, when some goal is reached at each step, the plans’ updates are needed constantly (i.e., at each time step). Thus, one may think of MAPF as a sequential decision-making problem -- at each time step, the following action (for all agents) should be decided. We assume the \emph{goal assignment} unit is external to the system, and the agents' behavior does not affect the goal assignments. We also assume that any LMAPF instance is additionally characterized by the \emph{episode length}, $L$, measured in time steps. After $L$ time steps have passed, the instance is considered to be done (despite some agents being on their way to the currently assigned goals).

Conventional MAPF success measures like $SOC$ or \emph{makespan} are not directly applicable to LMAPF. The most commonly used performance measure in LMAPF is the \emph{throughput} which is the average number of goals the agents achieve per one-time step. Technically it is computed as the ratio of the episode length to the total number of the reached goals.

\paragraph{Multi-agent Partially Observable Markov Decision Process} In our work, agents receive information about other agents not on the entire map but only in some local observation of their current position. We assume that each agent is aware of the global goals of other agents visible to them at the current moment. Additionally, each agent is assumed to possess a complete map of static obstacles. The observation function can be defined differently depending on the type of graph. In our experiments, we use 4-connected grids and assume that an agent observes the other agents in the area of the size $m\times m$, centered at the agent's current position. 

In such conditions of partial observability, the agent learns a policy function that allows it to generate a specific action by the observation. This setting can formally be represented as a partially observable multi-agent Markov decision process~\cite{bernstein2002complexity,Pack1998}: $M=\left\langle S, A, U, P, R, O, \gamma\right\rangle$.
At each timestep, each agent $u \in U$, where $U = {1, \dots, K}$, chooses an action $a_u \in A$, forming a joint action $\mathbf{j} \in \mathbf{J} = J^K$. This joint action leads to a change in the environment according to the transition function $P(s' | s, \mathbf{j}): S \times \mathbf{J} \times S \rightarrow [0, 1]$. After that, each agent receives individual observations $o_u \in O$ based on the global observation function $G(s, a): S \times A \rightarrow O$, whereas individual reward $R(s, u, \mathbf{j}): S \times U \times \mathbf{J} \rightarrow \mathbb{R}$, based on the current state, agent, and joint action. Thus, the joint reward is $r=\sum_u R(s, u, \mathbf{j})$. To make decisions, each agent conditions a stochastic policy by the observation $o^u$: $\pi_u(a_u | o_u): T \times A \rightarrow [0, 1]$. The task of the learning process is to optimize the policy $\pi_u$ for each agent to maximize the expected cumulative reward over time.

\paragraph{Monte-Carlo Tree Search} In our work, we use Monte-Carlo Tree Search (MCTS) as a model-based variant of the learnable agent's policy $\pi_u$. MCTS is a powerful  search method well-suited for sequential decision-making problems. Paired with state-of-the-art machine learning techniques, MCTS has recently achieved super-human performance in various board- and video games, see~\cite{silver2017mastering,ye2021mastering} for example. 

In MCTS-based methods, the agent picks an action given a state of the environment based on extensive simulating of how the environment would change and what rewards would be obtained if different actions are sequentially executed. MCTS is composed of four steps executed iteratively and intended to simultaneously build and explore the search tree: selection, expansion, simulation, and backpropagation. Selection is aimed at descending the constructed so far search tree. Conceptually, this can be seen as picking the most promising partial plan. To balance between the exploration and the exploitation, MCTS relies on assessing the nodes using the probabilistic upper confidence bound applied to the tree (PUCT)~\cite{rosin_multi-armed_2011}.

When the tree is descended, and the leaf node is picked, the latter is expanded by selecting an un-probed action and adding a new node to the tree. The added node is evaluated by simulating actions using a random or learnable policy, and the resulting reward is specially backpropagated through the tree. The process is repeated until the time budget is reached. When it happens, the action corresponding to the most visited outgoing edge of the root node is chosen to be executed. In this work, we will present our adaptation of MCTS for multi-agent partially-observable pathfinding.

\section{Method}

\begin{figure*}[htb!]
    \centering
    \includegraphics[width=1.0\textwidth]{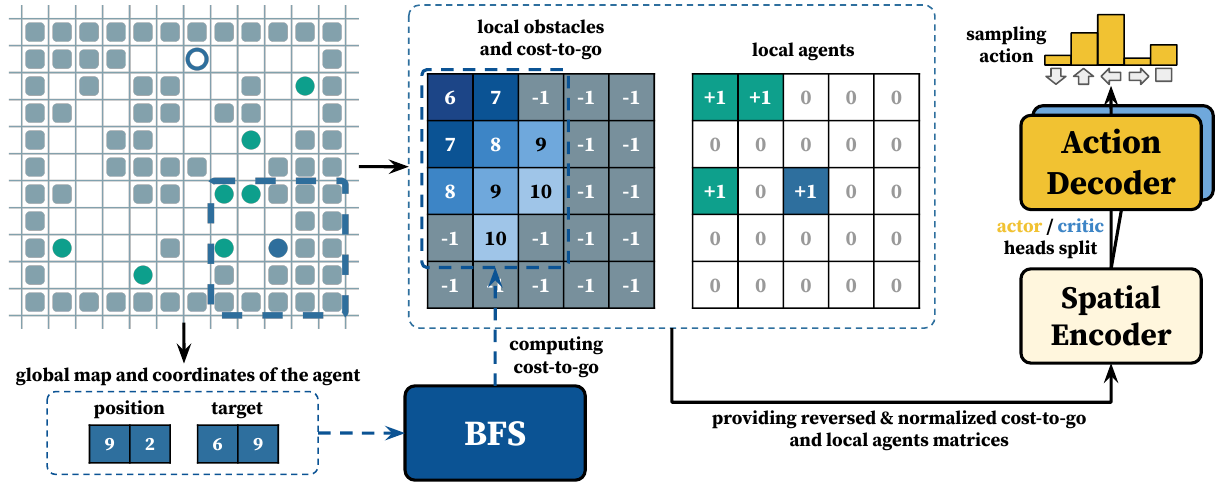}
    \caption{The figure depicts the scheme of the \textsc{CostTracer} algorithm. The approach  takes two matrices as input: one encodes obstacles, normalized reversed cost-to-go; the other has local agent positions. The entire pipeline is trained with the PPO algorithm, using a reward function that only provides positive feedback when the agent gets closer to its global goal.}
    \label{fig:scheme-cost-tracer}

\end{figure*}

Our method combines two principal ingredients. First, we employ the machinery of MCTS for an agent to reason about the possible future states of the environment and to choose the most promising action to be performed at the current time step, i.e., such action that, on the one hand, maximizes the chance of reaching the goal (eventually) and, on the other hand, decrease the chances of collisions and deadlocks with the other agents. Second, we use a learnable policy inside the MCTS simulation step. This policy is, indeed, approximated by a neural network and is tailored to accomplish MAPF tasks from the perspective of the single agent. We utilize the prominent actor-critic reinforcement learning method, i.e., Proximal Policy Optimization (PPO)~\cite{schulman2017proximal}, to pre-train such a policy. Importantly, as this policy is extensively used in MCTS to simulate the future states of the environment, it should be computationally efficient (fast). In practice, this means that the neural network that approximates the policy should contain a low number of parameters (weights). Motivated by this, we use a relatively compact neural network in this work that contains 161 thousand parameters compared to millions of them in conventional state-of-the-art learnable policies (e.g., the number of parameters in one of the recent methods we compare, SCRIMP, is about 9 million).

\subsection{Solving Decentralized MAPF Tasks with RL}

Numerous multi-agent Reinforcement Learning (MARL) algorithms can be used to solve the MAPF problem in partial observability. For incorporating an algorithm within MCTS in our case, the family of actor-critic methods, such as PPO~\cite{schulman2017proximal}, MAPPO~\cite{yu2022surprising}, or FACMAC~\cite{peng2021facmac}, is the most suitable. In our experiments, we utilize the PPO algorithm, which learns a shared policy independently for each agent.

In addition to choosing the algorithm, defining the observation space and reward function with which the algorithm will be trained in the environment is necessary. We employ design available local information comparable to one used in the Primal2 algorithm, meaning that the agent has information about static obstacles on the entire map, knows its current target, and can obtain information about other agents and their current targets in its field of view. 

We refer to this proposed approach as \textsc{CostTracer}, which emphasizes the design of the reward function and the neural network inputs. It utilizes only two input matrices and a simple reward function. The schematic representation of \textsc{CostTracer} is outlined in Figure~\ref{fig:scheme-cost-tracer}.

The agent's observation is defined as two matrices of the observation space size $m\times m$. 
The first matrix represents the positions of other agents ($+1$ if an agent is present and $0$ if not). The second matrix represents the normalized inverted cost-to-go function. Each time a target is received, the cost-to-go function is calculated using the breadth-first search (BFS) algorithm. It is provided to the agent in a normalized and inverted form. That is, a value of $1$ in the matrix corresponds to the closest cell to the target visible within the agent's observation. Obstacles are represented by $-1$, and all other values fall from $0$ to $1$.

We define the reward function as follows: the agent receives a reward of $+r$ if it reaches a cell closer to the goal on his current episode history. This information is  measured by the shortest distance using the cost-to-go function. In all other cases, the agent receives a reward of $0$. This reward function provides a dense signal while preventing exploitation of the reward function, as the agent's behavior that maximizes the reward guarantees getting close to the goal. 

Our neural network architecture employs a \textit{Spatial Encoder} and \textit{Action/Value Decoder} heads for both the actor and critic components, drawing inspiration from the AlphaZero approach~\cite{silver2017mastering} (see Figure~\ref{fig:scheme-cost-tracer}). The proposed architecture stands out by utilizing significantly fewer parameters than Primal2 and SCRIMP, enabling the algorithm to be trained on a single GPU in less than one hour. 
Despite its simplicity compared to other state-of-the-art algorithms, this setup demonstrates promising results, as shown in the experimental section.

\subsection{Multi-agent Neural MCTS for Intrinsic MDPs}

\begin{figure*}[ht!]
    \centering
    \includegraphics[width=1.0\textwidth]{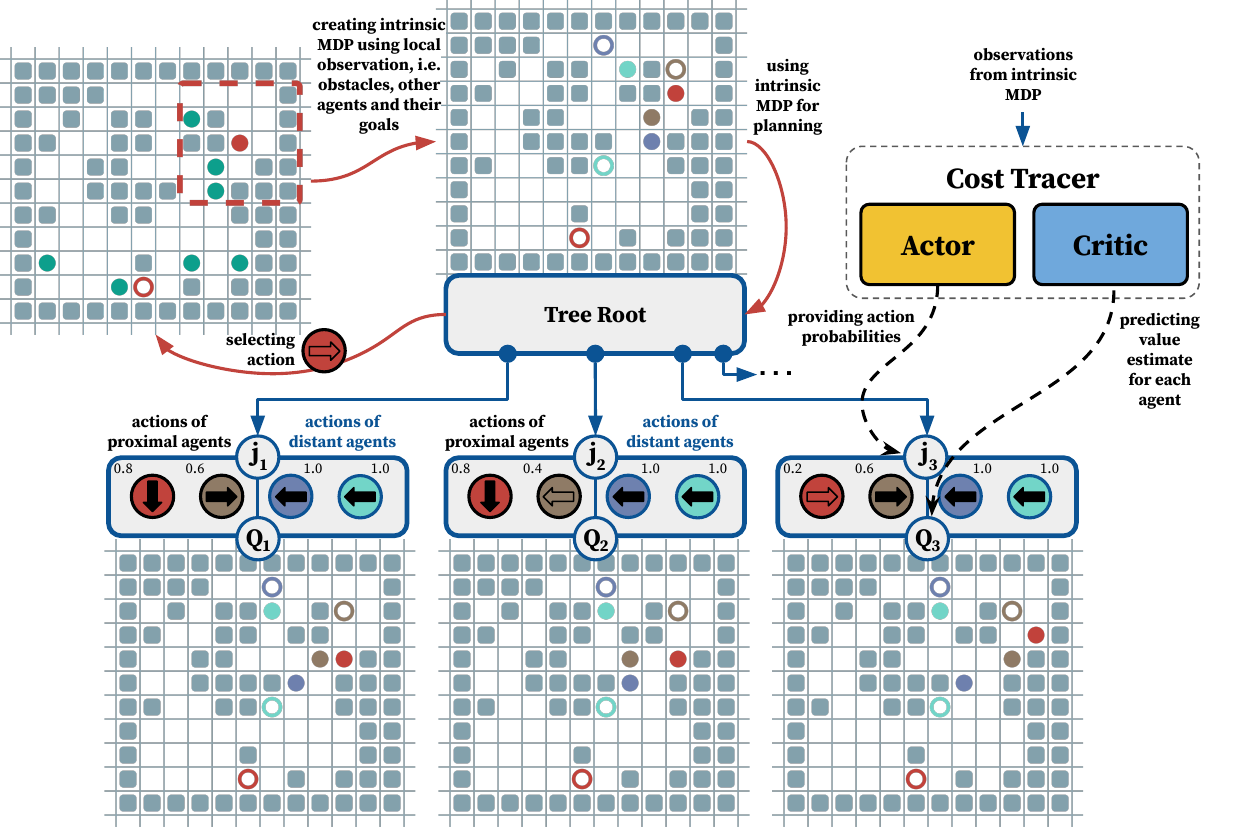}
    \caption{
The scheme of the \textsc{MATS-LP} approach. First, an intrinsic MDP (IMDP) is constructed using a global map of static obstacles, agents within the field of view, and their current targets. This MDP serves as the basis for planning using MCTS. In this approach, each tree node represents a joint action of all agents present in the IMDP, along with the associated node statistics. Action probabilities and node values are computed using the \textsc{CostTracer} algorithm. Notably, each agent's value is computed individually, and the node's estimate is derived as the sum of values across all agents.
The planning procedure exclusively concentrates on agents within close proximity. For example, in the given scenario, this includes the red agent itself and the brown agent. This deliberate decision to limit consideration to nearby agents significantly simplifies the decision-making process. Agent actions are visually depicted using circular representations, and bold arrows highlight actions featuring the highest probabilities. For distant agents, only a single action with maximum probability is considered.
}
    \label{fig:mmcts}

\end{figure*}

The scheme of multi-agent neural MCTS is sketched in Figure~\ref{fig:mmcts}.
Due to the fact that only partial information is available to each agent in the environment, the use of a centralized scheduler is not possible. In order to be able to plan in such situations, we suggest using \textit{intrinsic MDP} (IMDP). To do this, an intrinsic environment is created based on the egocentric observation of the agent (obstacles, other agents, and their current goals). Only the agents that the agent observes at the current step are included in this environment. All other cells that are not obstacles are considered empty.

Even within this intrinsic environment, the count of agents can be substantial. To tackle this, we present an action masking technique contingent on the proximity of other agents relative to the agent for which the planning is being conducted. The agents' proximity is established utilizing the BFS algorithm within their field of view. For the first $K$ agents, encompassing the agent itself, all feasible actions that avoid obstacles are contemplated (invalid actions are masked). Conversely, for the remaining agents, we adopt a greedy policy whereby only the action with the highest probability is used.
We denote the set of distant agents as $\mathbf{D}$ and restrict the action space of these agents to a single action with the highest probability, denoted as $A_\mathbf{D}^u = \argmax_{a_u \in A} \pi(o_u, a_u)$ (predicted by \textsc{CostTracer}). The final number of transitions is determined by multiplying the unmasked actions for each proximal agent.

During the lookahead search in such an MDP, the joint reward of $r$ for all agents is maximized. The reward function of the IMDP is identical to the reward function of  \textsc{CostTracer}.
Each node within the search tree corresponds to an intrinsic state $s$ of the IMDP. For every joint action $\mathbf{j}$ from state $s$, an edge $(s, \mathbf{j})$ is established to store a set of statistics $\{N(s,\mathbf{j}), Q, r, \pi_\mathbf{j}\}$. Here, $N$ represents the node visitation count, $Q$ is the mean joint Q-value, $r$ is the joint reward acquired from the IMDP upon executing action $\mathbf{j}$, and $\pi_j$ stands for the probability of joint action $\mathbf{j}$. Notably, we use the term $s$ to refer to the state of the IMDP.

The search process is divided into three distinct stages:

\paragraph{Selection.} Node selection is started from the Tree Root $s^0$, which is the initial state of the IMDP. The selection process continues until a leaf node is reached, which we denote as $s^l$, where $l$ represents the length of a single iteration of lookahead search. Each action is chosen based on the statistics stored in the nodes. This procedure follows PUCT bound, as utilized in the \cite{schrittwieser2020mastering}: 

\begin{equation*}
    \mathbf{j}^k = \argmax_{\mathbf{j}} \Bigg( {Q(s, \mathbf{j}}) + c\:\pi_j\,\frac{\sqrt{\sum_\mathbf{i}{N(s, \mathbf{i})}}}{1 + N(s,\mathbf{j})} \Bigg).
\end{equation*}

Here, $\mathbf{i}$ represents all possible joint actions from the current node, and $\pi_j=\prod_u{\pi_u(s,\mathbf{j})}$ is the probability of joint action. The constant $c$ controls the influence of the policy distribution on $Q$. Transition to the next state of the intrinsic environment is proceeded by applying $\mathbf{j}$ in it. $\pi_j$ and value estimate of the node $v^l = \sum_u{v(o_u, \mathbf{j})}$ is calculated using \textsc{CostTracer} and  the reward $r$ is accumulated using signal provided by IMDP. 

\paragraph{Expansion.} At the final timestep $l$, a new node is created. The transition to the next state of the intrinsic environment is carried out by applying $\mathbf{j}^l$ action. Action probabilities $\pi_u(s^l,\mathbf{j}^l)$ are calculated using \textsc{CostTracer}, and the reward for each agent $R(s,u, \mathbf{j})$ is accumulated using the signal provided by the  IMDP. The statistics of the new node are initialized as follows: $N^l(s^l, \mathbf{j}^l)=0, Q^l=0, r=\sum_u{R(s,u, \mathbf{j})}, \pi_j^l= \prod_u{\pi_u^l(o_u^l,a_u^l)}$.

\paragraph{Backpropagation.} This is the final step where accumulated statistics along the trajectory are updated. The update is computed using a discount factor $\gamma$, similar to the classic RL setup. To form an estimate of the cumulative discounted reward for the trajectory, we use:
\begin{equation*}
G^k = \sum_{\tau=0}^{l-1-k}\gamma^\tau r_{k+1+\tau}+\gamma^{l-k}v^l.    
\end{equation*}
After that the statistics for each edge $(s^{k-1}, \mathbf{j}^k)$ is updated as follows:
\begin{equation*}
\begin{aligned}
  Q(s^{k-1},\mathbf{j}^k) := &\: \frac{N(s^{k-1},  \mathbf{j}^k) + Q(s^{k-1},\mathbf{j}^k) + G^k}{N(s^{k},  \mathbf{j}^k) + 1}, \\ 
N(s^{k-1}, \mathbf{j}^k) := &\: N(s^{k-1},  \mathbf{j}^k) + 1.  
  \end{aligned}
\end{equation*}

The final action for the agent is determined as the action belonging to the most explored edge from the tree's root, determined by the number of visits \(N(s, \mathbf{j})\). The action \(a_u\) of the agent on behalf of which the IMDP was built is taken from \(\mathbf{j}\). The final joint action in the global environment is taken as the actions from all egocentric agents, planned with MCTS in their IMDPs. After executing this action in the environment, each agent receives their local observations, recreates its IMDP, and the process repeats.

\section{Empirical Evaluation}
\subsection{Experimental Setup}

To evaluate the efficiency of \textsc{MATS-LP}, we have conducted a set of experiments, comparing it with existing learnable approaches tailored to solve LMAPF problems. The episode length was set to $512$ in all the experiments. All the agents had the same parameters: their field-of-view was $11 \times 11$, all possible actions were considered only for the closest $3$ agents, including the main agent, $\gamma$-value was set to $0.96$, the number of expansions per iteration -- $250$, coefficient $c$ was set to $4.4$. More details and the values of the rest parameters are given in the Hyperparameters section below.

For the comparison, there were chosen two other learnable approaches -- a state-of-the-art method for solving LifeLong MAPF -- PRIMAL2~\cite{damani2021primal} and a recently presented method that has shown impressive results in solving single-shot MAPF -- SCRIMP~\cite{wang_scrimp_2023}. According to the results, presented in the original paper about SCRIMP, it clearly outperforms some other existing approaches -- PICO~\cite{Li2022MultiAgentPF} and DHC~\cite{ma2021distributed}. Thus, they were not taken as baselines.

We have used the implementation and the weights of the network provided by the authors of PRIMAL2\footnote{https://github.com/marmotlab/PRIMAL2} and SCRIMP\footnote{https://github.com/marmotlab/SCRIMP}. The code of SCRIMP has been adapted to solving LifeLong MAPF. To be more precise, the SCRIMP-local version was used, which has a limited communication radius (5) and shows better results. The size of the field-of-view for SCRIMP was set to $3\times3$, while for PRIMAL2 -- $11\times11$.

The comparison was conducted on three types of maps with different topologies. The first one consists of $20\times20$ grids with randomly placed obstacles. The density of obstacles varies from $0\%$ to $30\%$. In total 40 random maps were used. For each map 5 different instances with randomly placed start and goal locations were generated. The second type of map is the maze-like environments, that were generated using the generator taken from the PRIMAL2 repository. We have generated mazes with $10\times10$, $20\times20$, and $30\times30$ sizes, 50 maps per each size, 1 (randomly generated) problem instance per each map. Finally, the $33 \times 46$ warehouse map from~\cite{li2021lifelong} was used for evaluation. 10 random instances on this map were generated and used for evaluation.

To train the \textsc{CostTracer} algorithm, we used an open-sourced asynchronous implementation of the PPO algorithm\footnote{https://github.com/alex-petrenko/sample-factory}. A ResNet encoder as the \textit{Spatial Encoder} with one residual layer was utilized, and the hidden layer sizes for the multi-layer perception (MLP) blocks were set to $32$ for the \textit{Action/Value Decoder}. The training process had a discount factor ($\gamma$) of $0.96$ and a learning rate of $0.00019$. More detailed parameter descriptions can be found in the Hyperperameters section below. 
We employed a Bayesian hyperparameter search to optimize the algorithm's parameters and architecture. In total, we conducted $100$ algorithm runs, which roughly corresponds to $120$ GPU hours using a single Titan RTX. The model that showed the best results with fewer parameters was chosen. 

\subsection{Results}

\begin{figure*}[htb!]

    \centering
    \includegraphics[width=0.76\textwidth]{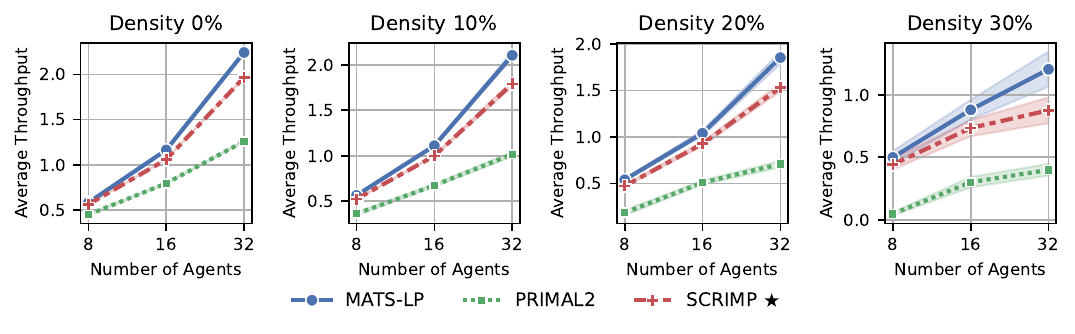}
    \hspace{5px}
    \includegraphics[width=0.18\linewidth]{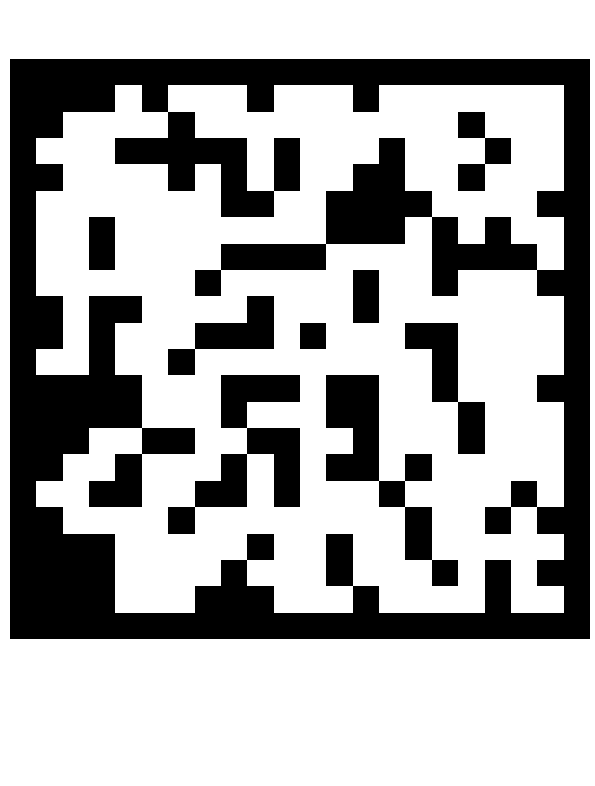}
    
    \caption{Average throughput of MATS-LP, SCRIMP, and PRIMAL2 on random maps $20\times20$ with various obstacle densities. The $\star$ symbol marks the approaches that were trained on the corresponding type of maps. The shaded areas indicate the 95\% confidence intervals.}
    \label{fig:results_random}

\end{figure*}

\begin{figure*}[htb!]

    \centering
    \includegraphics[width=0.7\linewidth]{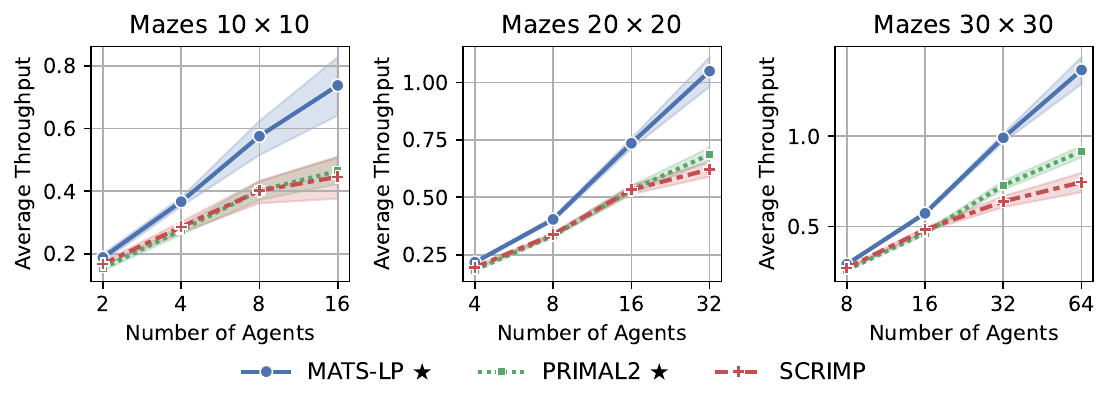}
    \hspace{15px}
    \includegraphics[width=0.2\linewidth]{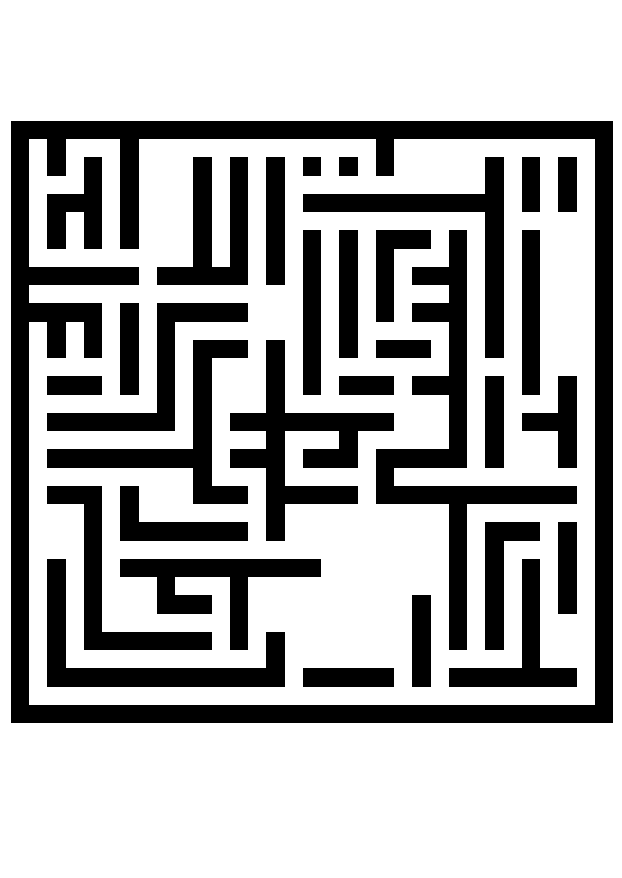}
    \caption{Average throughput of MATS-LP, SCRIMP and PRIMAL2 on maze-like maps with various sizes. The $\star$ symbol marks the approaches that were trained on the corresponding type of maps.}
    \label{fig:results_mazes}

\end{figure*}

The results on the random maps are presented in Figure~\ref{fig:results_random}. 
In this experiment MATS-LP outperforms SCRIMP in all the cases, gaining $15.6\%$ higher throughput on average. At the same time, PRIMAL2 demonstrates poor performance with more than twice less throughput than MATS-LP on average. Such behavior is explained by the fact, that PRIMAL2 is tailored to solve maps that consist of corridors, such as mazes environments. Moreover, it was trained on maze-like maps, similar to MATS-LP. Thus, this type of map is out-of-distribution for these two approaches.
The shaded areas indicate the 95\% confidence intervals. A detailed analysis of the results has shown that throughput can vary significantly from map to map, as some maps contain a narrow passage dividing the map into two parts, and many agents get stuck trying to pass through the passage in opposite directions, blocking each other.

The results of the second series of experiments on maze-like maps of various sizes are presented in Figure~\ref{fig:results_mazes}. As well as in the first series of experiments, MATS-LP significantly outperforms both competitors. Compared to SCRIMP it has shown $46.2\%$ higher throughput on average, while PRIMAL2 was outperformed by $38.8\%$. 
In most of the cases SCRIMP and PRIMAL2 demonstrate almost the same efficiency on average with only exception of $30\times30$ maze maps with $32$ or $64$ agents where PRIMAL2 substantially outperformed SCRIMP demonstrating a bit better scalability on such type of maps.

The last series of experiments involved a warehouse-like that was taken from \cite{li2021lifelong}. We utilized the same way of generating start and goal locations for the agents as in the original paper, when start locations for all the agents might be placed only on the left or right edge of the map, while goal locations - only near the obstacles in the middle of the map. Due to the limitations imposed to the possible start locations, the total amount of agents cannot exceed number of 192. Following these rules, we have generated 10 different problem instances.

\begin{figure*}[t!]

    \centering
    \includegraphics[width=0.24\textwidth]{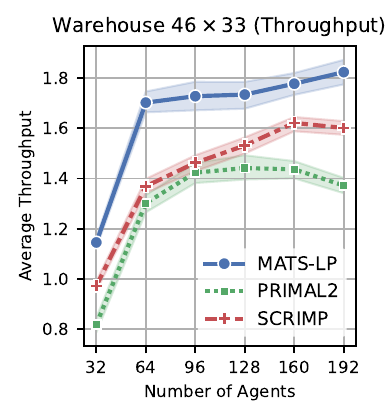}
    \includegraphics[width=0.24\textwidth]{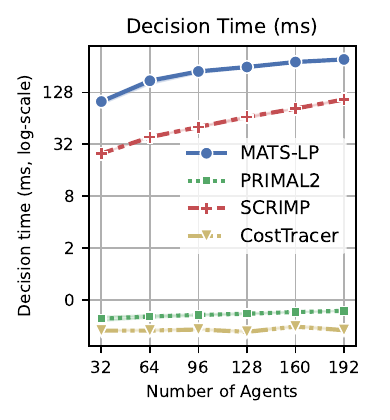}
    \includegraphics[width=0.24\textwidth]{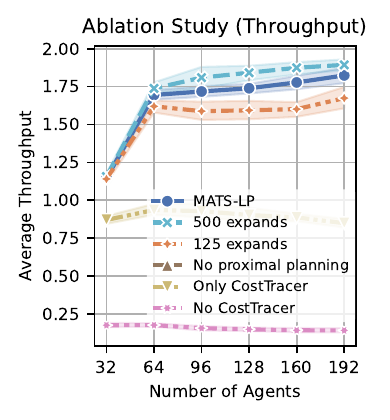}
    \hspace{10px}
    \includegraphics[width=0.24\textwidth]{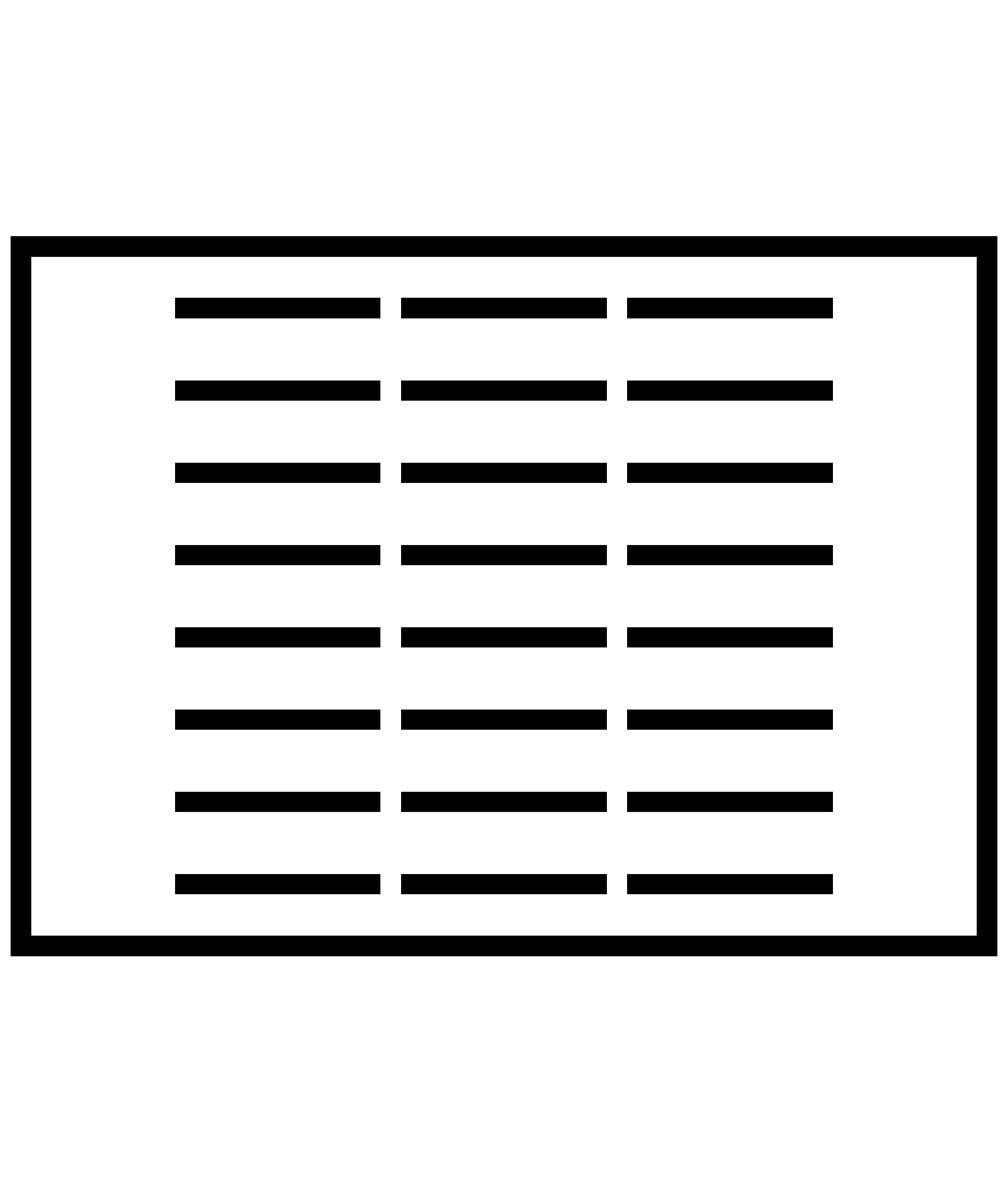}
    
    \caption{Average throughput and average decision time of MATS-LP, SCRIMP and PRIMAL2 and ablation study of MATS-LP on warehouse-like map. The shaded areas indicate the 95\% confidence intervals.}
    \label{fig:warehouse}

\end{figure*}

The results of these experiments are shown in Figure~\ref{fig:warehouse}. In addition to measuring the average throughput of the approaches, we have also estimated the time required to make a decision about the next action per each agent and conducted the ablation study of MATS-LP. 

The left plot of Figure~\ref{fig:warehouse} shows the averaged throughput. Again, MATS-LP demonstrates better performance, its throughput is $15.8\%$ higher than the one of SCRIMP (on average), and $27.1\%$ higher then the one of PRIMAL2.

The middle plot demonstrates the time required by each of the solvers to make a decision about the next action for a single agent (decision time). We added the line for the \textsc{CostTracer}, the learnable policy used within MATS-LP, to this plot. Clearly its decision time is almost constant, while the one of MATS-LP increases from 103ms to 300ms when the number of agents goes from 32 to 192. This may be explained by increasing number of agents that appear in the field-of-view of each agent and the fact that MATS-LP predicts the actions of these observable agents, which takes time (as it necessitates running \textsc{CostTracer} more frequently). Similarly, the decision time for SCRIMP is not constant and ranges from 25ms to 107ms, due to the need for coordinating movements with a larger number of agents. Although MATS-LP requires more time than SCRIMP to make a decision, its scalability is slightly better. I.e., the difference in decision time between 192 agents and 32 agents is 3x for MATS-LP and 4x for SCRIMP.

The right plot in Figure~\ref{fig:warehouse} demonstrates the results of the ablation study for MATS-LP. \textsc{CostTracer} is MATS-LP with MCTS turned off. We have also evaluated MATS-LP with a random policy instead of \textsc{CostTracer}. The term ``No proximal planning" refers to the variant where planning is executed solely for an egocentric agent, selecting only actions with highest probability for other agents. Furthermore, we experimented with increasing the number of expansions to $500$ and reducing them to $125$, compared to the $250$ expansions used by the basic version of MATS-LP. The worst results are demonstrated by the version that utilizes random policy instead of \textsc{CostTracer}, that indicates its crucial importance. Next lowest throughput is obtained by \textsc{CostTracer}, whose throughput is almost twice worse compared to MATS-LP. The results of the version, that is planning for an egocentric agent only, get worse with increasing number of agents as the increase of density of agents increases the need of coordination between them. The versions with increased/decreased number of expansions show results slightly better or worse than the basic version respectively. The latter indicates, that while MATS-LP has a relatively high decision time, it actually can be adjusted to the required decision time or even work in anytime fashion, adapting to a specific time budget.

\subsection{Hyperparameters}

Table~\ref{table:parameters} presents the hyperparameters of \textsc{CostTracer} and \textsc{MATS-LP} approaches. ``Number of agents" denotes the number of agents in the environment in which \textsc{CostTracer} was trained. The table's parameters marked ``tuned" were optimized using Bayesian search. We used default values commonly used in other studies for the other parameters. The parameter root exploration ratio corresponds to the noise (with uniform distribution) added in the tree root that facilitates exploration in the MCTS algorithm.

\begin{table}[ht!]
    \caption{Parameters of \textsc{CostTracer} and \textsc{MATS-LP}. }
    \centering
    \label{table:parameters}
    \begin{tabular}{clc}
        \toprule
        \textsc{CostTracer} & Value & Tuned \\
        \midrule
        \midrule
        
        Adam learning rate & \num{0.00013} & \checkmark \\
        $\gamma$ (discount factor) & $0.96$ & \checkmark\\
        PPO clip ratio &  $0.2$ \\
        PPO optimization epochs & $1$ & \checkmark \\ 
        Batch size & $2048$ & \checkmark\\
        Entropy coefficient & \num{0.06878} & \checkmark\\
        GAE$_\lambda$ & $0.95$ \\
        \midrule
        ResNet residual blocks & $2$ & \checkmark\\ 
        ResNet number of filters  & $32$ & \checkmark\\
        Activation function       & ReLU \\
        Network initialization & orthogonal \\
        MLP size & $32$ & \checkmark\\ 
        Number of agents & [64, 128] & \checkmark \\ 
        Parallel environments & $16$ \\
        Training steps & \num{75000000} \\
        Observation patch & $11\times11$ \\
        \midrule
        Network parameters & \num{161734} \\ 
        \bottomrule
        \\
        \toprule
        \textsc{MATS-LP} & Value & Tuned \\
        \midrule
        \midrule
        Discount factor $\gamma$ & $0.96$ \\ 
        Exploration coefficient $c$ & $4.4$ & \checkmark\\
        Number of expansions & $250$ & \\
        Planning agents $K$ & $3$ & \checkmark\\
        Root exploration ratio & $0.6$ & \checkmark \\
        \bottomrule
    \end{tabular}
    
\end{table}

\section{Conclusion}
In this work, we have studied the Lifelong MAPF problem and suggested a solver based on the Monte-Carlo Tree Search equipped with the (lightweight) learnable policy tailored to solve MAPF from the individual agent's prospective. The resultant solver is decentralized and does not require explicit agent communication. Empirically we have shown that our solver can generalize well to the unseen LMAPF instances and outperform the state-of-the-art competitors in different challenging setups. A prominent direction for future research is to develop a fully learnable MCTS for LMAPF, i.e., to learn the simulation policy with MCTS itself like in~\cite{schrittwieser2020mastering}.

\bibliography{lib}

\end{document}